\documentclass{article}
\usepackage{ijcai23}

\usepackage{times}
\usepackage{soul}
\usepackage{url}
\usepackage[hidelinks]{hyperref}
\usepackage[utf8]{inputenc}
\usepackage[small]{caption}
\usepackage{graphicx}
\usepackage{amsmath}
\usepackage{amsthm}
\usepackage{booktabs}
\usepackage{algorithm}
\usepackage{algorithmic}
\usepackage[switch]{lineno}

\usepackage{newfloat}
\usepackage{listings}
\usepackage{multirow}
\usepackage{amssymb}
\usepackage[table,xcdraw]{xcolor}

\usepackage[american]{babel}
\usepackage{microtype}

\title{Single-stage Multi-human Parsing via Point Sets and Center-based Offsets}

\author{
Jiaming Chu$^{1\dag}$
\and
Lei Jin$^{1*\dag}$\and
Junliang Xing$^2$\And
Jian Zhao$^{3}$\footnote{Lei Jin and Jian Zhao are corresponding authors. $^\dag$ Jiaming Chu and Lei Jin are equal contribution. }
\affiliations
$^1$Beijing University of Posts and telecommunications \\
$^2$Tsinghua University\\
$^{3}$Institute of North Electronic Equipment\\
\emails
\{chujiaming886, jinlei\}@bupt.edu.cn,
jlxing@tsinghua.edu.cn,
zhaojian90@u.nus.edu 
}

\begin{document}

\maketitle

\begin{abstract}
% xing: v2
This work studies the multi-human parsing problem. Existing methods, either following top-down or bottom-up two-stage paradigms, usually involve expensive computational costs. We instead present a high-performance \textbf{S}ingle-stage \textbf{M}ulti-human \textbf{P}arsing (SMP) deep architecture that decouples the multi-human parsing problem into two fine-grained sub-problems, \textit{i.e.,} locating the human body and parts. SMP leverages the point features in the barycenter positions to obtain their segmentation and then generates a series of offsets from the barycenter of the human body to the barycenters of parts, thus performing human body and parts matching without the grouping process. Within the SMP architecture, we propose a \emph{Refined Feature Retain} module to extract the global feature of instances through generated mask attention and a \emph{Mask of Interest Reclassify} module as a trainable plug-in module to refine the classification results with the predicted segmentation. Extensive experiments on the \emph{MHPv2.0} dataset demonstrate the best effectiveness and efficiency of the proposed method, surpassing the state-of-the-art method by 2.1\% in AP$_{50}^p$, 1.0\% in AP$_{vol}^p$, and 1.2\% in PCP$_{50}$. In particular, the proposed method requires fewer training epochs and a less complex model architecture. We will release our source codes, pretrained models, and online demos to facilitate further studies.
\end{abstract}

\begin{figure}[t]
\centering
\includegraphics[width=1.0\columnwidth]{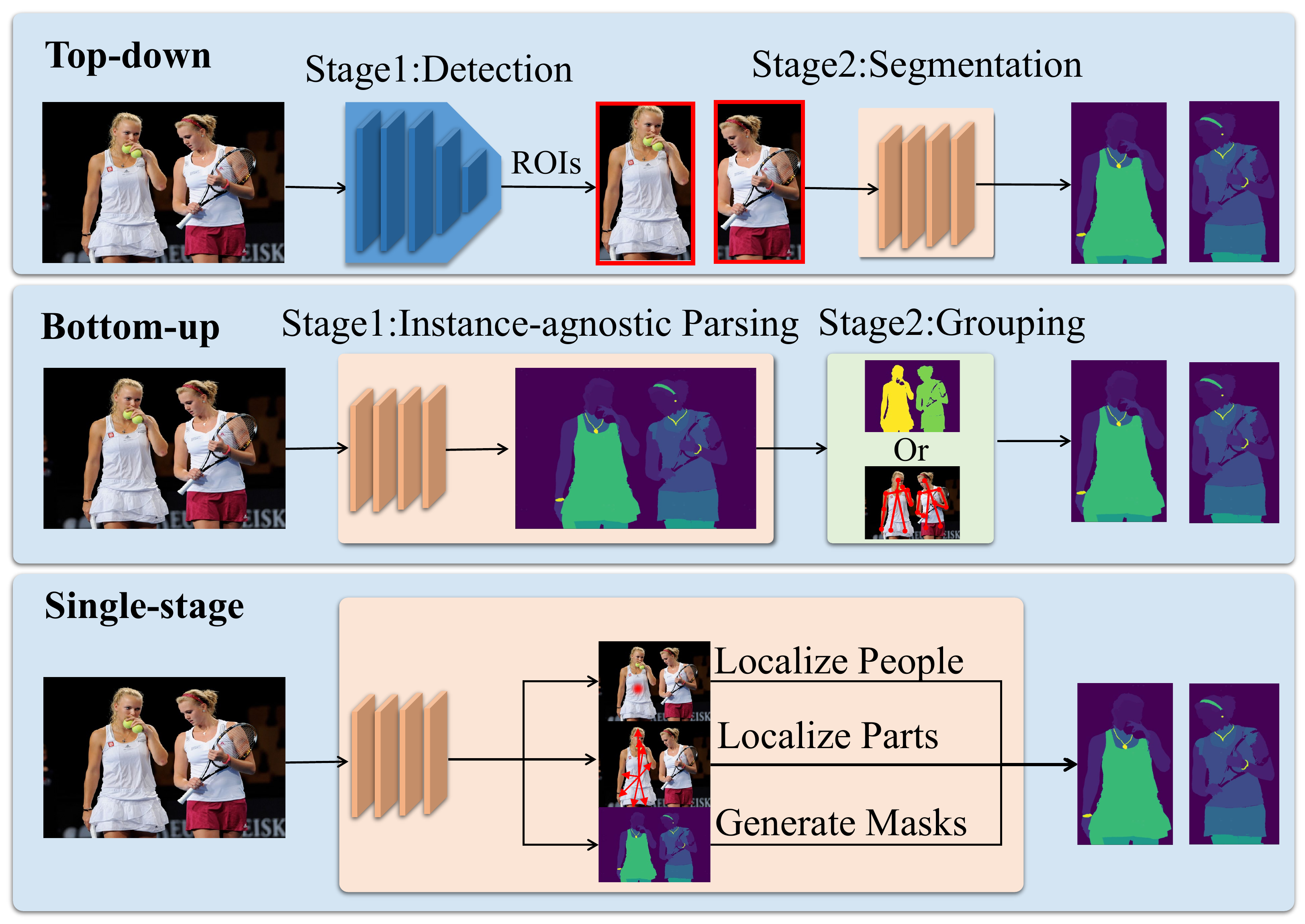} % Reduce the figure size so that it is slightly narrower than the column. Don't use precise values for figure width.This setup will avoid overfull boxes.
\caption{The differences between top-down, bottom-up, and single-stage methods in multi-human parsing task. The single-stage method is parallel computing, and neither the instance detection of the top-down method nor the grouping process of the bottom-up method are required. Best viewed in color.}
\label{Differences between frameworks}
\vspace{-3mm}
\end{figure}

\vspace{-3mm}

\section{Introduction}
%先介绍问题，再讲以前的解决方法，再讲自己的方法。
Instance-Aware Multi-Human Parsing (IAMHP) aims to cut apart the parts of human body according to semantics, and group them by human instance. Compared with semantic segmentation task and instance segmentation task, it is more challenging.
Because for each pixel in the picture, it is necessary to judge not only the semantic label of the part level, but also the instance label of the human level. 
This paper attempts to propose a superior framework to conduct instance-aware multi-human parsing, which can significantly simplify the complicated processing pipelines of previous researches.

%Semantic and instance segmentation tasks ~\cite{AlexanderKirillov2022PanopticS} aim to predict instance label and semantic label for each pixel in the image and have wide applications ~\cite{XueqingDeng2022NightLabAD, FanMingyuan2021RethinkingBF, hu2021learning}. However, these tasks only consider the human instance level understanding or semantic understanding, where the fine-grained human part perception is under-explored. This information is also essential for several customized applications, $e.g.$, virtual try-on, virtual game.
%To this end, Instance-Aware Multi-human Parsing (IAMHP)~\cite{JianZhao2018UnderstandingHI,JianZhao2017SelfSupervisedNA} has been introduced recently, which aims to judge the semantic label of the part level as well as the instance label of the human level for each pixel in a given image.

A few efforts have been proposed to model the IAMHP problem. The existing multi-human parsing method could be roughly divided into two categories: bottom-up and top-down methods. The bottom-up methods~\cite{JianZhao2017SelfSupervisedNA} usually parse all the human parts through instance segmentation or semantic segmentation, and then group the human parts belonging to the same human instance through the different grouping methods, $e.g.$, clustering by human instance segmentation, instance edge prediction, $etc$. The top-down methods~\cite{LuYang2018ParsingRF,LuYang2020RenovatingPR} usually detect human instances first, and then parse human parts for each human instance individually. Albeit encouraging results achieved, these methods still have the following limitations. Both bottom-up and top-down methods are involved two stages and complicated post-processings, $i.e.$, part parsing and grouping, instance detection and single-person parsing. Such sequential stages lead to redundant processes and high computational cost. 

More recently, to eliminate redundant processes and improve model calculation efficiency, researchers proposed single-stage framework~\cite{XuechengNie2019SingleStageMP,ZigangGeng2021BottomUpHP,JinLei2022SingleStageIE,WilliamMcNally2021RethinkingKR,TianfeiZhou2021DifferentiableMH} for the human pose estimation task and have achieved impressive results. The key idea of single-stage framework is a holistic representation for the human instance, $e.g.$, the hierarchical structure pose representation in SPM~\cite{XuechengNie2019SingleStageMP} and the decoupled 3D pose representation in DRM~\cite{JinLei2022SingleStageIE}. Overall, these methods only need a center point and center-based offsets to represent the human pose. The simple model structure and parallel inference schema greatly improve the efficiency of model inference. However, for multi-human parsing task, only center point can not encode fine-grained semantic information for human parts. To construct a single-stage multi-human parsing framework, how to unify human-instance-level information and human-part-level information is still a challenging and unexplored problem.

In this paper, we explore the possibility to understand human body with point sets and center-based offsets to conduct multi-human parsing. Specifically, the point sets are composed of human barycenter and the barycenters of parts, and the center-based offsets are the displacements from human barycenter to part barycenters. With such representation, we realize a Single-stage Multi-human Parsing (SMP) framework, which omits the time-consuming grouping process. In particular, we decouple the IAMHP task into four sub-tasks, termed, human instance localization, part instance localization, part instance segmentation and subordination mapping prediction of the two granularity instances. %We use the full convolution structure to generate the barycenter heatmaps of the instances, representing the instances with points. And we generate fine mask by point features for part instance. The offset vectors from the barycenter of human to that of the parts is regressed to represent the mapping relationship. %if it is the barycenter of the human instance. 
Additionally, we also propose the Refined Feature Retain (RFR) module and the Mask of Interest Reclassification (MIR) module. The former utilizes the instance correlation between mask features as an attention to make the model learn more relevant features. The latter can refine the classification results by taking the output of part segmentation as region of interest.

%Compared with top-down and bottom-up methods, the proposed SMP has superior simplicity and efficiency.
Without the instance detection in top-down methods and the complex grouping process in bottom-up methods, our method completes the prediction of mapping relationship in parallel while predicting the location and segmentation of instances, thus improving the inference efficiency. Specifically, our method reduces the inference time by 102.5\% and 91.2\% with ResNet-50~\cite{KaimingHe2015DeepRL} and ResNet-101 as backbone, respectively. %is significantly faster than all bottom-up methods in inference speed, and about 1 time faster than the fastest top-down method. 
Finally, it is worth mentioning that, SMP also achieves state-of-the-art performance, outperforming the best result~\cite{ZhangfuDong2022MNetR2} by 2.1\% in AP$_{50}^p$, 1.0\% in AP$_{vol}^p$, and 1.2\% in PCP$_{50}$ on MHPv2.0~\cite{JianZhao2018UnderstandingHI} dataset with the same settings. To show the generalization of SMP, we test SMP on DensePose COCO~\cite{RizaAlpGuler2018DensePoseDH} (Please refer to the appendix). Experiments show that SMP achieves superior performance on DensePose COCO as well. %Our source codes, pretrained models, and online demos will be released upon acceptance.

Our contributions are summarized as follows.

\begin{itemize}
    \item We make the first attempt to understand human body with point sets and center-based offsets. With this as a guideline, to the best knowledge of us, we introduce the first single-stage framework for instance-aware multi-human parsing.
    \item We propose a Refined Feature Retain (RFR) module which uses mask attention to guide the model to extract features, and further put forward a Mask of Interest Reclassification (MIR) module to optimize the classification results of the model.
    \item Our method significantly outperforms all the state-of-the-art top-down and bottom-up methods in AP and PCP metrics on MHPv2.0 benchmark, obtaining fastest inference speed.
\end{itemize}

\begin{figure*}[t]
\centering
\includegraphics[width=2.1\columnwidth]{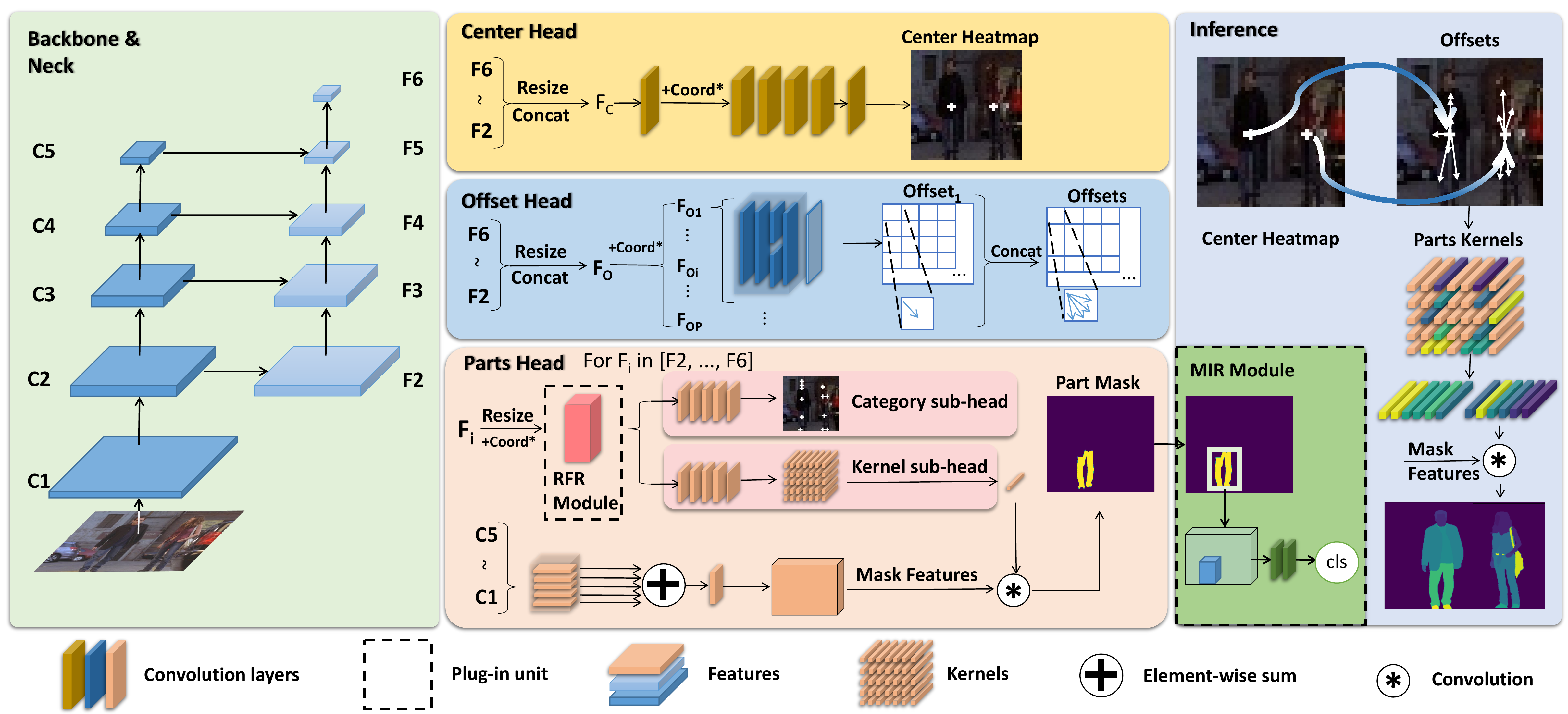} 
\caption{The Illustration of our Single-stage Multi-human Parsing (SMP) framework for instance-aware human semantic parsing. Coord* refers to concatenating the relative coordinates to the feature. Details can be seen in Sec.~3.1 and Sec.~4.1. Best viewed in color.}
\label{Overview}
\end{figure*}

\vspace{-3mm}

\section{Related Work}

\subsection{Instance Segmentation}

In general, existing instance segmentation techniques can be roughly categorized as top-down or bottom-up methods.

As a typical top-down method, Mask R-CNN~\cite{KaimingHe2017MaskR} first detects instances, and then performs semantic segmentation to achieve pixel-level predictions. Dai et al. ~\cite{JifengDai2016InstancesensitiveFC} utilize the location sensitivity of instances to segment instances based on the FCN network~\cite{JonathanLong2014FullyCN}, and assign the prediction task of instance segmentation to different grids. 

On the other hand, the bottom-up methods aim to learn better pixel-level feature representations. Some of them adopt the loss function about clustering~\cite{DavyNeven2019InstanceSB,HuiYing2021EmbedMaskEC}, and others attempt to utilize grids to generate strongly related location features~\cite{XinlongWang2019SOLOSO,DanielBolya2019YOLACTRI}. But both of them output intensive predictions for each grid or pixel, which incurs redundant calculation, as not all pixels belong to specific instances. 

In order to reduce unnecessary computation, a handful of works attempt to leverage conditional convolution~\cite{BrandonYang2019CondConvCP}. It replaces the conventional convolution kernel with the predicted features, further reducing considerable computation amount. So there are also many works~\cite{ZhiTian2020ConditionalCF,KonstantinSofiiuk2019AdaptISAI,TeroKarras2018ASG,XinlongWang2020SOLOv2DA} like Condinst, SOLOv2, utilize dynamically generated kernel to improve the capacity of the model, hence the mask head can be more compact, leading to faster inference speed. Such similiar kernel mechanism is also widely used in other fields~\cite{GaofengCao2022KPNMFIAK,ZehuiChen2023AutoAlignPF,ZhenLiangNi2020BARNetBA}.

\subsection{Multi-human Parsing}
%\smallskip\noindent\textbf{Multi-Human Parsing}
Compared with instance segmentation~\cite{ZhangfuDong2022MNetR2}, semantic segmentation~\cite{HexinDong2022RegionAwareML,WenbinHe2022SelfsupervisedSS,JieLi2021IMENetJ3} and pose estimation~\cite{ZihaoZhang2021Sequential3H}, multi-human parsing is a task that not only needs to distinguish the part instance label of each pixel, but also needs to determine the subordinate relationship between human parts and human instances. Similarly, existing multi-human parsing methods also can be categorized into top-down or bottom-up methods. 

The top-down methods usually need detector and feature aligning. Parsing RCNN~\cite{LuYang2018ParsingRF} and RP-RCNN~\cite{LuYang2020RenovatingPR} both generate bounding box first and utilize a geometric and context encoding (GCE) module or GSE-FPN to get finest semantic pixel features. 
In addition, some methods~\cite{TaoRuan2019DevilIT} utilize extra ground truth to enhance the ability of the model to extract instance context features. SNT~\cite{RuyiJi2019LearningSN} which relies on Mask RCNN to distinguish human instances, predicts human parts by attributes in stages, reducing the difficulty of prediction. 

The bottom-up methods do not give priority to distinguishing human instances. They usually predict the instance label and semantic label of pixels and group them. Some works~\cite{JianZhao2020FineGrainedMP,JianshuLi2017TowardsRW} apply the GAN-based network to improve the learning ability to instances and semantic features of the model. Other works utilize additional groundtruth information such as pose~\cite{TianfeiZhou2021DifferentiableMH} and edge~\cite{KeGong2018InstancelevelHP} to help the model learn instance characteristics.

\vspace{-2mm}

\section{Method}
% 先讲模型总体架构
%The key idea of the proposed single-stage framework is to simplify the process in multi-human parsing task, improving the efficiency. As shown in Fig. 2, three dedicated heads and two modules are utilized to achieve instance-aware multi-human parsing. Here, we first introduce the problem formulation and details of different heads, then we present our proposed RFR module and MIR module.

\noindent\textbf{Problem Formulation.} Given an image $I$, multi-human parsing aims to parse the human instances in $I$, and obtain their masks of body parts $\mathcal{M} = {\{M^{human}_{m}\}}^{N}_{m=1}$ , where $N$ denotes the total number of human instances in $I$. For each human instance, it is composed of part instance segmentation  $M^{human}_{m} = {\textstyle \sum_{i=1}^{C_P} M^{part}_i }$, where $C_P$ is the number of part categories, $M^{part}_i$ is the mask of $i$-th class part.

%\subsection{Single-stage Multi-human Parsing}

\subsection{Overview}
The overview of our Single-stage Multi-human Parsing (SMP) framework is illustrated in Fig.~\ref{Overview}. Firstly, we send an image to the Features Pyramid Network (FPN)~
\cite{TsungYiLin2016FeaturePN} to generate feature maps with different sizes. Then we process the feature maps with three dedicated heads, \emph{i.e.}, center head, offset head, parts head, to predict human location and mask information. Finally, we can obtain instance-aware multi-human parsing results with the outputs of the three heads.
%The core of the single-stage method mainly lies in three structurally decoupled and functionally complementary heads, including center head which localizes the positions of human instances, part head which localizes the positions of part instances and generates masks of parts, and offset head which regresses the mapping relations between parts and human.

\noindent\textbf{Center Head.} The center head aims to predict the location of each individual human instance. To avoid the overlapped center problem, we utilize the barycenter of visible mask to represent each instance. Given an image $I$, the backbone and FPN extract its multi-scale semantic features $F_{i} (i \in {2,...,6})$, as seen in Fig.~\ref{Overview}. Next, We resize the multi-scale features to the same size $S \times S$ with bilinear interpolation ($S$ is set as 40), then concatenate them, and apply a $3 \times 3$ convolution layer for feature fusion and compression. The fusion feature $F_C$ can simultaneously perceive human instances of multiple scales. To further enhance the position information of the features, we adopt the coordinate convolution (CoordConv)~\cite{RosanneLiu2018AnIF} to concatenate the relative position with the fusion features. %Further, four consecutive convolution layers are used for local semantic feature extraction. 
Finally, we transfer the features by convolution layers to a one-channel heatmap $H_{Center}^{S \times S \times 1}$. The values in the $H_{Center}^{S \times S \times 1}$ represent the existing confidence of the center.

\noindent\textbf{Offset Head.} 
%The mapping relationship can be conceptualized as $O^{N_{ins} \times 2 \times C_{P}}$, where $N_{ins}$ is the number of human instances, $2$ refers to the dimensions of the mapping vectors, and $C_{P}$ is the number of part classes. In this paper, we adopt center-based offsets to capture the mapping relationship. 
In this paper, we predict the offsets from the barycenter of human body to the barycenters of its corresponding part instances to estimate the mapping relationship.
To get features ($F_O$ in the offset head of Fig.~\ref{Overview}) from different scales, similar to the center head, we apply bilinear interpolation to resize the features to $S \times S$ and concatenate them together (S is set as 40), and utilize CoordConv~\cite{RosanneLiu2018AnIF} to obtain location information. Next, We employ the decoupled structure with AdaptBlock~\cite{ZigangGeng2021BottomUpHP} to predict the offset map for each human part.
%The structure of decoupled branch is composed of AdaptBlock~\cite{ZigangGeng2021BottomUpHP,XizhouZhu2018DeformableCV} and a convolution layer. 
The input of each decoupled branch is only one part of the fusion features, which greatly reduces computation cost. All decoupled branch outputs will eventually be concatenated together to obtain offset maps $M_{Offset}^{S \times S  \times C_{P} \times 2}$, where $C_P$ is the number of parts class. The pixel around the human barycenter in offset maps contains $C_{P}$ regression vectors, pointing to the barycenter positions of the $C_{P}$ parts from the pixel.

\begin{table}[t]
\renewcommand\arraystretch{1.1}
\caption{The settings of feature grids. Instances in different scales response in different feature maps. Human refers to the grids applied in center head and offset head.}
\setlength{\tabcolsep}{3.5pt}
\small
\begin{tabular}{lcccccc}
\hline
\textbf{}    & \textbf{} & \textbf{} & \textbf{Parts} & \textbf{} & \textbf{} & \textbf{Human} \\ \hline
\textbf{Level}   &$F_2$   &$F_3$   &$F_4$    &$F_5$  &$F_6$  & fusion           \\
\textbf{Scale}   &$\leq 96$   &(48,192)  &(96,384)   &(192,768) &\textbf{$\geq 384$} &\textbf{-}           \\
\textbf{Grids}         & 40      & 36      & 24     & 16    & 12  & 40  \\ \hline             
\end{tabular}
\captionsetup{font={small}}
\label{The numbers of feature grids setting}
\end{table}

\noindent\textbf{Part Head.} 
The goal of part head is to localize the part instances and generate their fine-grained masks. Inspired by SOLOv2~\cite{XinlongWang2020SOLOv2DA} and YOLOv3~\cite{JosephRedmon2018YOLOv3AI}, we utilize Features Pyramid Network (FPN) to predict part instances with different scales. The outputs of each level are resized to different grid sizes and compressed to the same number of channels. Instances with different scales are corresponding to different levels as shown in Tab.~\ref{The numbers of feature grids setting}. The outputs are utilized as input for each prediction head: parts category sub-head and parts kernel sub-head. The parameters of these heads are shared across different levels. We adopt a similar model like center head, and we also use the barycenter of visible part mask to represent each part. A convolution layer with $C_{P}$ output channels is at the end of the \textbf{part category sub-head}, and we could obtain the parts heatmap $H_{Part}^{S \times S \times C_{P}}$ finally. 

After localizing part instance, we refer to condition convolution~\cite{ZhiTian2020ConditionalCF} and SOLOv2 to generate the refined mask. The input feature is also the single-layer feature of FPN, and the size is aligned with the $H_{Part}$. This sub-head also concatenates the relative coordinates by CoordConv. Its structure is the same as that of the $H_{Part}$ sub-head, but the number of channels in the last convolution layer is $C_K$ which is set as 256, then the final output of \textbf{part kernel sub-head} is $F_{Kernel}^{S \times S \times C_{K}}$. 

To ensure the fineness of the generated mask, the feature map will not be downsampled to $S \times S$. The features employ the multi-scale output features of the backbone (C1,...,C5 in Fig.~\ref{Overview}). The features at different levels are all upsampled to $1 / 4$ the original resolution through bilinear interpolation. With element-wise summation and a convolution layer, the convoluted feature map $F_{Mask}^{h \times w \times 256}$ is obtained, where $h, w$ are $1/4$ height and width of the original image.

\begin{figure}[t]
\centering
\includegraphics[width=1\columnwidth]{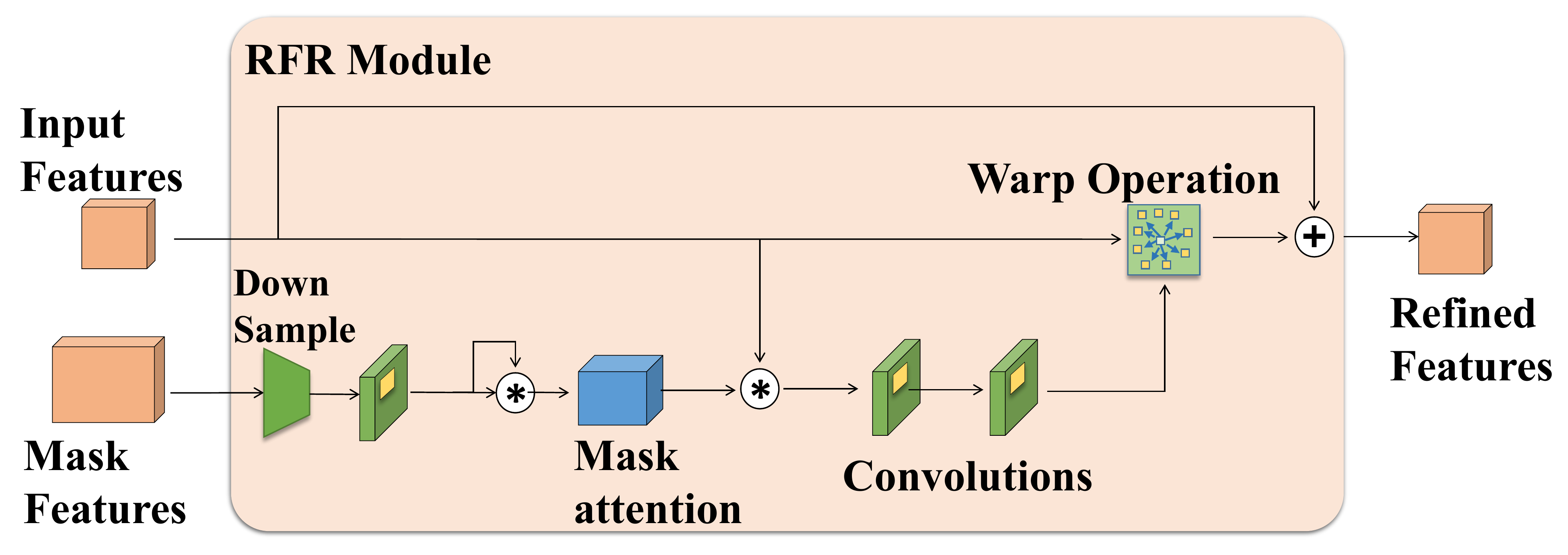} % Reduce the figure size so that it is slightly narrower than the column. Don't use precise values for figure width.This setup will avoid overfull boxes.
\caption{The detailed architecture about RFR module. It is placed after downsampling of category and kernel branch as shown in Fig.~\ref{Overview}.}
\label{RFR}
\end{figure}

\begin{figure*}[h]
\centering
\includegraphics[width=2\columnwidth]{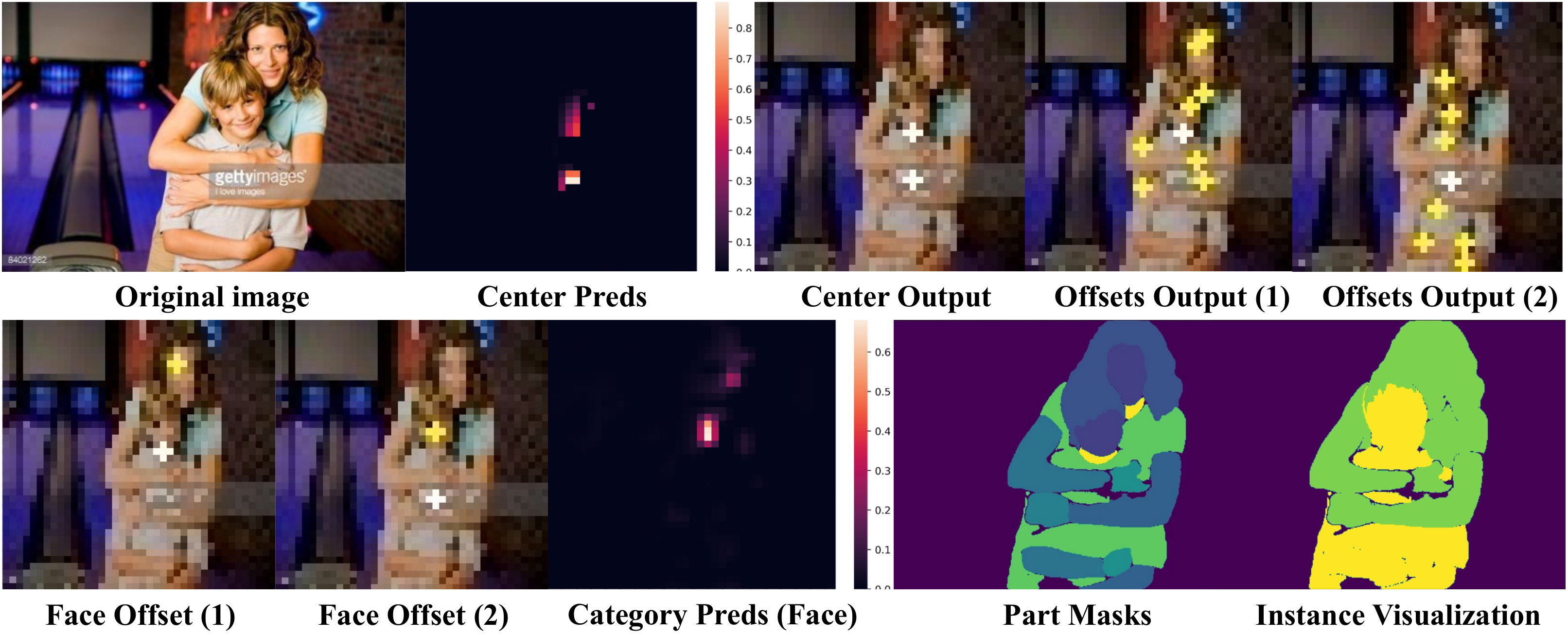} % Reduce the figure size so that it is slightly narrower than the column. Don't use precise values for figure width.This setup will avoid overfull boxes.
\caption{The visualization results of intermediate outputs for SMP. Center preds and Category preds (Face) are the visual confidence maps (“preds” is short for “predictions”). Center output is generated by maxpooling layer based on center preds. Yellow crosses in offsets output are the part instance locations pointed by offsets and white crosses are human centers. (1), (2) denote different human instances. Except original image, parts mask and instance visualization, other visualizations are resized to $S \times S$ (S=40~during~inference) as description in Sec.~3.1 and Sec.~4.1. Best viewed in color.}
\label{Interresults}
\vspace{-0.8em}
\end{figure*}

\subsection{Refined Feature Retain Module}
To guide the model to better extract the features around the barycenter, we propose Refined Feature Retain (RFR) module. The position of RFR module in our framework SMP is shown in Fig.~\ref{Overview}.

The main idea of RFR module is utilizing the mask feature as an attention to guide the learning of the category branch. Part head completes the instance segmentation by conditional convolution. For each value on output segmentation results, it is actually the inner product of the convolution kernel and the corresponding feature on the feature map. When the vectors are nearly normalized, the inner product can be regarded as the cosine similarity between them. During training, the projection between the positive convolution kernel and the convolution features on the positive position will gradually increase. Since the prediction label is limited to $[0,1]$, the convolution result will be closer to the cosine similarity. As seen in Fig.~\ref{RFR}, through the self-correlation calculation of the convoluted feature map, we could get the attention map of the instance on the corresponding position. The self-attention map, namely mask attention, has superior instance guidance. By multiplying the category feature with the mask attention of each position, we could obtain a new refined feature map. We utilize the new feature as offset input and conduct warp operation to guide model adaptively obtain more instance information.

\subsection{Mask of Interest Reclassify Module}
The RFR module enhances the ability of feature extraction in the category sub-branch. However, the model is still difficult to classify the fashion classes with strong inter-class similarity~\cite{ZhengboZhang2022DistillingID}. For the classification task, the model prefers to learn the target with relatively stable scale. Therefore, we propose a Mask of Interest Reclassify (MIR) module that can be trained quickly, and it can be used as a plug-in module to improve the model performance. The position of MIR module in our framework SMP is shown in Fig.~\ref{Overview}.%This module can be added to the trained model, and a few short epochs of training can take effect. 

%Mask-RCNN relies on region of interest (ROI) to achieve scale invariance of classification tasks. 
Our model can output fine-grained segmentation as the Region of Interest (ROI) to achieve secondary classification. MIR module is separate and could utilize the output results of other branches. We choose the fusion feature of FPN as the input feature, extract the feature through consecutive convolution layers, and use semantic segmentation labels to supervise the feature. The mask generated by the part head is used as ROI to obtain local features. The features are reduced to a fixed size through the ROIalign with the size of 14, and the convolution layer with the kernel size of 14 is employed to compress the features again. Finally, two consecutive full connection layers are applied to output the classification results.

% 损失函数
\subsection{Loss Computation}
The output of our model is divided into three parts: the heatmaps of the barycenters of both the human body and the parts, the mapping relationships between these two types of barycenter, and the fine masks of the parts. For each type of outputs, we apply distinct labels for supervision. 

\noindent\textbf{Center and Part Localization Loss.} For the heatmaps localizing the barycenter of two granularity instances, we employ the traditional Focal Loss~\cite{TsungYiLin2017FocalLF}. Given the barycenter $(c_{x}, c_{y})$ and the height $h$ and width $w$ of the instance, the barycenter is extended to a rectangular area called center region $(c_{x}, c_{y}, \epsilon h ,\epsilon w)$, where $\epsilon$ is a hyperparameter to control the scale of region ($\epsilon$ is set as 0.2). Any grid that falls into the center region is regarded as a positive sample. %Due to the part instances with different scales corresponding to feature with different sizes, every instance generates about 6 positive samples.

\noindent\textbf{Offset Loss.} The offset prediction can be viewed as a regression problem. The supervision labels of offset head are the offset vectors. The offset vectors on each grid are obtained by subtracting the grid coordinates from the barycenter coordinates of the corresponding parts.
\begin{equation}
\small
    Offset_{i,j,p} = (c_{px} - i,c_{py} - j),
\end{equation}
where $i,j \in [0,S-1]$ are the grid coordinates, $c_{px}$ and $c_{px}$ are the barycenter coordinates of the person's part, $p \in  [0,C_p-1]$ is the label of part category. We adopted Smooth $\ell_1$ Loss~\cite{RossGirshick2015FastR} for the regression problem. When calculating the loss, we will multiply each grid with a scale-related weight to balance the impact of scale.

\noindent\textbf{Mask Loss.} For the segmentation of part instances, the label supervision method is the same as the general instance segmentation. We apply Dice Loss~\cite{hcr:FaustoMilletari2016VNetFC} for optimization.

\noindent\textbf{Total.} The loss of SMP is the sum of the loss for each head.
\begin{equation}
\small
\mathcal{L}_{\rm total}  = \lambda_{c} \mathcal{L}_{\rm center} +\lambda_{p} \mathcal{L}_{\rm part}  
                 +\lambda_{d} \mathcal{L}_{\rm dice} +\lambda_{o} \mathcal{L}_{\rm offset},
\end{equation}
where $\lambda_{c}$ and $\lambda_{p}$ are both used to balance the center and part loss, with a value of 1.0. $\lambda_{d}$ is used to balance the mask loss, with value of 3.0. $\lambda_{o}$ is used to balance the Smooth $\ell_1$ Loss for offsets, with value of 10.0. The above parameters are verified by the experiment search and SMP gets better results.

\begin{table*}[t]
\renewcommand\arraystretch{1.1}
\caption{The comparable results on MHPv2.0 dataset. $*$ denotes the backbone with DCNv2~\protect\cite{XizhouZhu2018DeformableCV}. For better comparison, we mark the state-of-the-art methods under the same settings in gray. RP-RCNN with 150 epochs is the best available pretrained model which is also used in the qualitative comparison (RP-RCNN is inferior than AIParsing with ResNet-50 and 75-epoch training strategy).}
\vspace{-2mm}
\setlength{\tabcolsep}{14.5pt}
\small
\begin{tabular}{llcccccc}
\hline
Methods      & Backbone     & Extra data & \multicolumn{1}{c|}{Epoch}   & AP$^p_{50}$ & AP$^p_{vol}$ & PCP$_{50}$  \\ \hline
Top-Down     &              &            &                               &      &       &      \\ \hline
M-RCNN~(He et al. TPAMI'17.)       & ResNet-50     & -          & \multicolumn{1}{c|}{-}        & 14.9 & 33.9  & 25.1 \\
P-RCNN~(Yang et al. CVPR'18.)      & ResNeXt-101      & Pose       & \multicolumn{1}{c|}{75}      & 30.2 & 41.8  & 44.2 \\
M-CE2P~(Ruan et al. AAAI'19.)       & ResNet-101      & -       & \multicolumn{1}{c|}{150}      & 34.5 & 42.7  & 43.7 \\
SNT~(Ji et al. ECCV'19.)       & ResNet-101         & -          & \multicolumn{1}{c|}{-}        & 34.4 & 42.5  & 43.5 \\
RP-RCNN~(Yang et al. ECCV'20.)  & ResNet-50    & Pose       & \multicolumn{1}{c|}{75}      & 40.5 & 45.2  & 39.2 \\
RP-RCNN~(Yang et al. ECCV'20.)  & ResNet-50    & Pose       & \multicolumn{1}{c|}{150}      & 45.3 & 46.8  & 43.8 \\ 
AIParsing~(Zhang et al. TIP'22.)  & ResNet-50      & -       & \multicolumn{1}{c|}{75}      & 41.1 & 45.9  & 45.3 \\
\rowcolor[HTML]{E6E6E6} AIParsing~(Zhang et al. TIP'22.)  & ResNet-101      & -       & \multicolumn{1}{c|}{75}      & 43.2 & 46.6  & 47.3 \\ \hline
Bottom-up    &            &                               &      &       &      \\ \hline
PGN~(Gong et al. ECCV'18.)   & ResNet-101  & -       & \multicolumn{1}{c|}{-}        & 17.6 & 35.5  & 26.9 \\
MHParser~(Li et al. TOMM'21.) & ResNet-101    & -          & \multicolumn{1}{c|}{-}        & 18.0 & 36.1  & 27.0 \\
NAN~(Zhao et al. IJCV'20.)  & -        & -          & \multicolumn{1}{c|}{$\sim$80} & 25.1 & 41.8  & 32.3 \\
\rowcolor[HTML]{E6E6E6} DSPF~(Zhou et al. CVPR'21.) & ResNet-101  & Pose       & \multicolumn{1}{c|}{150}      & 39.0 & 44.3  & 42.3 \\ \hline
Single-stage &            &                               &      &       &      \\ \hline
SMP (Ours)   & ResNet-101  & -          & \multicolumn{1}{c|}{12}       & 38.4 & 44.8  & 43.0 \\
SMP (Ours) & ResNet-101$^*$   & -          & \multicolumn{1}{c|}{12}       & 43.0 & 46.4  & 45.5 \\
SMP (Ours)   & ResNet-50  & -          & \multicolumn{1}{c|}{36}       & 42.0 & 46.1  & 46.1 \\
SMP (Ours) & ResNet-50$^*$   & -          & \multicolumn{1}{c|}{36}       & 46.9 & 48.0  & 50.1 \\
\rowcolor[HTML]{E6E6E6} SMP (Ours) & ResNet-101   & -          & \multicolumn{1}{c|}{36}       & 45.3 & 47.6  & 48.5 \\
SMP (Ours)  & ResNet-101$^*$   & -          & \multicolumn{1}{c|}{36}       & \textbf{47.1} & \textbf{48.2}  & \textbf{51.5} \\ \hline
\end{tabular}
\captionsetup{font={small}}

\label{The main results on MHPv2.0}
\vspace{-1em}
\end{table*}

\vspace{-2mm}

\section{Experiment}
\subsection{Experiment Setup}

\noindent\textbf{Dataset.} MHPv2.0~\cite{JianZhao2018UnderstandingHI} is the largest and most challenging dataset in the field of multi-human parsing. It contains 15,403 training images, 5,000 validation images. Each picture contains 2 - 26 human instances, with an average of 3. The categories of 58 parts include 11 human body parts labels and 47 clothing and accessory labels. Meanwhile, the data set also provides labels for 16 keypoints of human pose. To show the generalization of SMP, we also conducted experiments on the DensePose COCO dataset~\cite{RizaAlpGuler2018DensePoseDH}. See the appendix for the relevant results.

\noindent\textbf{Metrics.}
For instance-aware human parsing performance, we employ the Average Precision based on part (AP$^{p}$)~\cite{JianZhao2018UnderstandingHI} for multi-human parsing evaluation, which calculates mean part instance level pixel IoU of different semantic part categories within a human instance to determine if the human instance is a true positive. We choose the AP$^{p}_{50}$ and AP$^{p}_{vol}$ as the evaluation metrics. The former defines the instance whose IoU is larger than the threshold of 0.5 as the positive, and the latter is the average of AP$^{p}$ in the IoU threshold ranging from 0.1 to 0.9 in increments of 0.1. In addition, we also report the official metric, Percentage of Correctly parsed semantic Parts (PCP)~\cite{JianZhao2018UnderstandingHI}.

\noindent\textbf{Implementation Details.}
We implement our SMP based on mmdetection~\cite{mmdetection} on a server with 8 NVIDIA Tesla V100 GPUs and 32GB memory per card. We adopt ResNet-50, ResNet-101 and FPN as backbone and neck in all architectures, each of which is trained end-to-end. A mini-batch involves 32 images. We use scale jitter where the shorter image side is randomly sampled from 640 to 800 pixels, following SOLO~\cite{XinlongWang2019SOLOSO}. We regard 12 epochs as 1$\times$ training with an initial learning rate of 0.02, which is then divided by 10 at 9-th and 11-th epoch. We use 3$\times$ training schedule whose learning rate is divided by 10 at 27-th and 33-th epoch for performance comparison. 

\noindent\textbf{Inference.}
Since this work is the first attempt that the single-stage method is applied to the instance-aware multi-human parsing task, we design an inference scheme dedicated to the single-stage method. Fig.\ref{Interresults} shows the visual intermediate outputs of SMP. Since SMP represents the instance parsing as point sets and center-based offsets, %which can be output in parallel in a single stage, our inference process does not need to perform grouping operations, $e.g.$, bipartite graph or Hungarian algorithm. 
we utilize maxpooling to remove the redundant human barycenters in $H_{Center}$, and obtain the offsets to get the positions $(C_x,C_y)$ of the part instances corresponding to the human instance. Before getting the positions of part instances by offset, we resize $H_{Part}$ and $F_{Kernel}$ to the same size as $M_{Offset}$. We judge whether the human instance contains the part instance by comparing the confidence value in $H_{Part}$ on the position pointed by the offset vectors with the threshold. And the redundant part instances will be filtered out. After localizing the part instances, We could the obtained their corresponding convolution kernel from $F_{Kernel}$, and the final parsing result can be obtained by convolution operation.

\begin{figure*}[t]
\centering
\includegraphics[width=2\columnwidth]{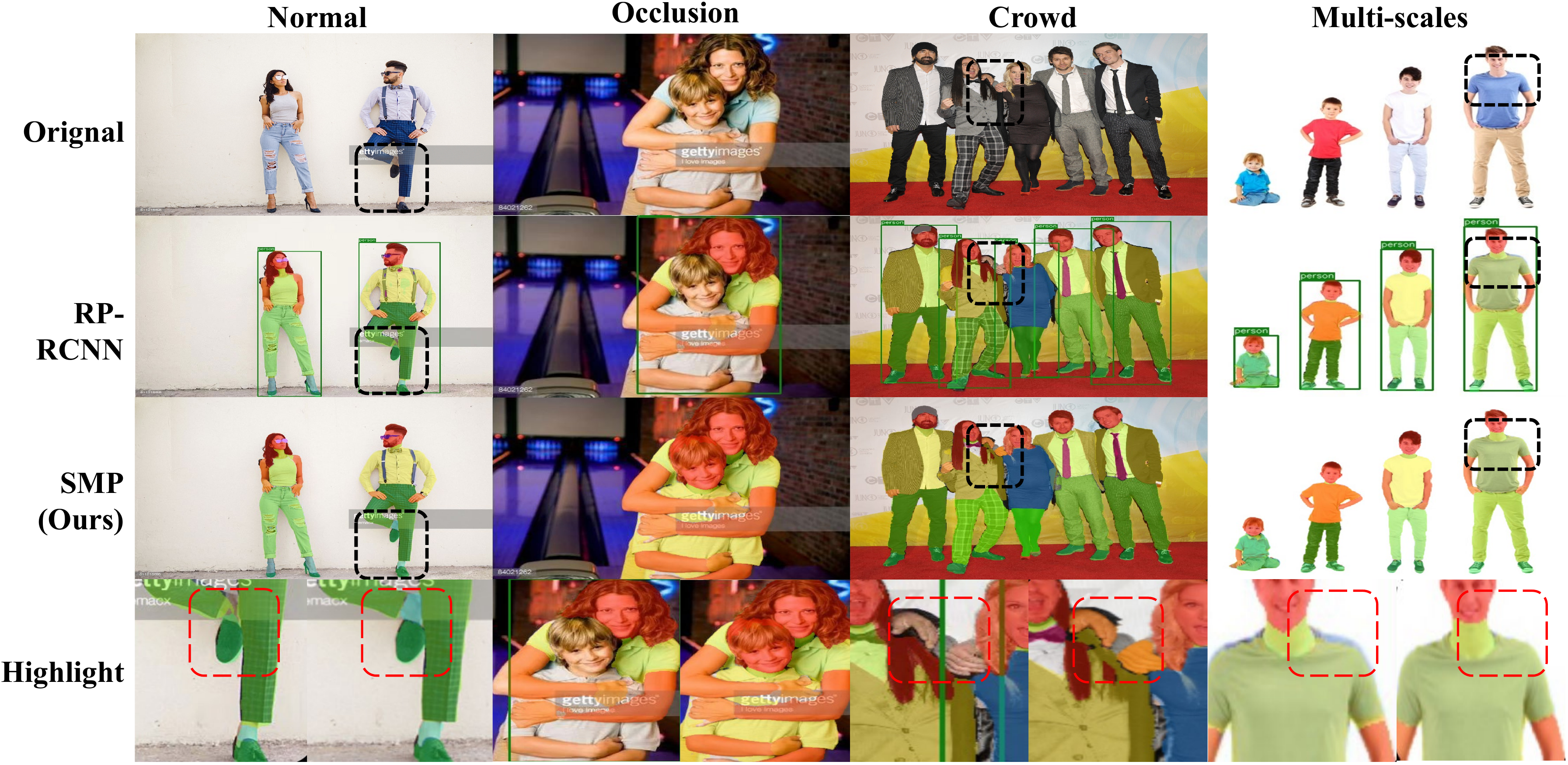} % Reduce the figure size so that it is slightly narrower than the column. Don't use precise values for figure width.This setup will avoid overfull boxes.
\caption{The visualization results from SMP including instances with different scales and quantity. In the highlight row, the left is RP-RCNN, the right is SMP. Best viewed in color.}
\vspace{-3mm}
\label{visual}
\end{figure*}

\subsection{Main Results}
\noindent\textbf{Comparison with State-of-the-art Models.} Tab.~\ref{The main results on MHPv2.0} reports the results of comparison with the state-of-the-art top-down and bottom-up methods on MHPv2.0 val set. Our method is significantly superior to the existing top-down and bottom-up model in all instance-aware metrics with less training epochs and without the aid of additional ground truth other than mask. For other methods, the longer learning schedule is set up to 150 epochs while our method sets 36 epochs as 3$\times$ learning schedule. With deformable convolution in backbone, SMP achieves 47.1\%, 48.2\% and 51.5\% on AP$^{p}_{50}$, AP$^{p}_{vol}$ and PCP$_{50}$. Based on ResNet-101, ours outperforms by +2.1, +1.0 and + 1.2 than AIParsing~\cite{SanyiZhang2022AIParsingAI} with less training time. Compared with DSPF~\cite{TianfeiZhou2021DifferentiableMH}, which is the SOTA model in bottom-up methods, SMP outperforms on all metrics by +6.3 on AP$^{p}_{50}$, +3.3 on AP$^{p}_{vol}$ and +6.2 on PCP$_{50}$.

\noindent\textbf{Qualitative Results.} As shown in Fig.~\ref{visual}, we visualize some comparison results with the top-down method RP-RCNN~\cite{LuYang2020RenovatingPR} (RP-RCNN performs best in the open-source works). We present four different images belonging to different scenarios: normal, occlusion, crowd, and multi-scales. In the normal and the multi-scales images, we can find that SMP performs better in the parsing of small parts, $e.g.$, half of the left foot, and the edge processing of the instance. For the occlusion image, RP-RCNN has an error detection, in which two people are detected as one. Crowd pictures are challenging for the multi-human parsing task. Too many people make it difficult for the part head respond and easy to arise miss detection. However, compared with RP-RCNN, SMP has no problem of ROI detection error, and its ability to distinguish categories is also superior.

\begin{table}[t]
\renewcommand\arraystretch{1.1}
\setlength{\tabcolsep}{8.5pt}
\caption{The results of ablative study on MHPv2.0 val set.}
\vspace{-2mm}
\small
\begin{tabular}{cc|ccc}
\hline
RFR          & MIR          & AP$^p_{50}$ & AP$^p_{vol}$ & PCP$_{50}$ \\ \hline
             &              & 42.5       & 45.8  & 44.1 \\
             & $\checkmark$ & 42.8 (+0.3) & 46.1 (+0.3) & 44.7 (+0.6) \\
$\checkmark$ &              & 42.6 (+0.1) & 46.3 (+0.5) & 44.3 (+0.2) \\
$\checkmark$ & $\checkmark$ & 43.0 (+0.5) & 46.4 (+0.6)  & 45.5 (+1.4)  \\ \hline
\end{tabular}
\captionsetup{font={small}}
\label{The results of ablation study on MHPv2.0 val.}
\end{table}

\subsection{Empirical Experiments}

\noindent\textbf{Ablative Experiments.} In this section, we performed ablative study on Refined Feature Retain module and Mask of Interest Reclassify module. The model in the ablative study was trained by the short learning schedule (the number of epochs is 12), and the backbone of the model is ResNet-101 with the deformable convolution~\cite{XizhouZhu2018DeformableCV}. From Tab.~\ref{The results of ablation study on MHPv2.0 val.}, we can see that the performance of the model is enhanced with the addition of RFR module and MIR module. From the results of ablation experiments, it can be seen that after adding RFR module to the model, all metrics have been improved. As a separate module added after the model, MIR module can be quickly trained which is trained with 6 epochs. We utilize the Focal Loss~\cite{TsungYiLin2017FocalLF} for supervision.

\begin{table}[t]
\renewcommand\arraystretch{1.2}
\caption{The runtime analysis on MHPv2.0 val set with average 2.6 people per image. The \textcolor{blue}{blue} values are the decreases in inference time compared with the fastest open-source reproducible model. $*$ denotes the backbone with DCNv2~\protect\cite{XizhouZhu2018DeformableCV}.}
\vspace{-2mm}
\setlength{\tabcolsep}{3.2pt}
\small
\begin{tabular}{l|lc}
\hline
\multicolumn{1}{l|}{}             & Methods  & Inference time (ms) \\ \hline
\multirow{4}{*}{Top-down}         & SNT~\cite{RuyiJi2019LearningSN}       & 3546           \\
                                  & M-CE2P~\cite{TaoRuan2019DevilIT}   & 1023           \\
                                  & P-RCNN~\cite{LuYang2018ParsingRF}    & 256           \\
                                  & RP-RCNN~\cite{LuYang2020RenovatingPR}   & 341           \\ \hline
\multirow{3}{*}{Bottom-up}        & MHParser~\cite{LuYang2020RenovatingPR}  & 1224           \\
                                  & NAN~\cite{JianZhao2018UnderstandingHI}       & 997           \\
                                  & PGN~\cite{KeGong2018InstancelevelHP}      & 524           \\
                                  \hline
\multirow{2}{*}{Single-stage}  & SMP (Ours, ResNet-50$^*$)     & 124 \textcolor{blue}{($\downarrow$ 102.5\%)}          \\
& SMP (Ours, ResNet-101$^*$)     & 132 \textcolor{blue}{($\downarrow$ 91.2\%)}           \\ \hline
\end{tabular}
\captionsetup{font={small}}
\label{The runtime analysis on MHPv2.0 val with average six people per image.}
\vspace{-1.2em}
\end{table}

\begin{table}[]
\renewcommand\arraystretch{1.1}
\caption{The impact of grid numbers and center region on MHPv2.0 val set. 1$\times$ denotes the grids setting which is the same as Tab.~\ref{The numbers of feature grids setting}. 0.5$\times$, 2$\times$ denote the grid numbers multiplied by the corresponding coefficients. $\epsilon$ is the hyperparameter controlling the scale of center region. FPS indicates the inference speed.}
\vspace{-2mm}
\setlength{\tabcolsep}{10.7pt}
\small
\begin{tabular}{cc|cccc}
\hline
Grids & $\epsilon$& AP$^{p}_{50}$   & AP$^{p}_{vol}$ & PCP$_{50}$ &FPS \\ \hline
0.5$\times$  & 0.2   & 26.6 & 36.7    & 31.8  &8.45  \\
0.5$\times$ & 0.1   & 25.7 & 36.3   & 29.9  &8.45  \\ \hline
 1$\times$    & 0.2   & 43.0 & 46.4    & 45.5  &7.47  \\
 1$\times$    & 0.1   & 41.9 & 45.8    & 44.0  &7.47  \\ \hline
 2$\times$    & 0.2   & 44.1 & 47.1    & 47.3  &5.94  \\
 2$\times$    & 0.1   & 44.0 & 47.1    & 46.2  &5.94  \\ \hline
\end{tabular}
\captionsetup{font={small}}
\label{grid experiments.}
\vspace{-1.3em}
\end{table}

\noindent\textbf{Running Time Analysis.} In addition, we also conducted runtime analysis with the open-source reproducible methods. The experiment is conducted on one V100 GPU. The existing top-down and bottom-up methods all take multi-stage paradigms, leading to computational redundancy. From Tab.~\ref{The runtime analysis on MHPv2.0 val with average six people per image.}, we can see that the speed of our method is far faster than the bottom-up methods, and is about 1 time faster than the fastest top-down method Parsing R-CNN~\cite{LuYang2018ParsingRF}. (SMP reduces the inference time by 102.5\% and 91.2\% with ResNet-50 and ResNet-101 respectively).

\noindent\textbf{Impact of Grid Numbers.}
We also conduct additional experiments about grids number and hyperparameter $\epsilon$ under the 12-epoch training schedule. The setting of grids number in Tab.~\ref{The numbers of feature grids setting} and $\epsilon=0.2$ can stably generate 2-6 positive labels for each instance. The above setting is effective for learning about semantic features of part head due to the consistency of instance scales. As the results in Tab.~\ref{grid experiments.}, under the grid numbers with the 2$\times$ size, the performance of SMP is improved by about 1\% in all indicators, but the running speed is reduced by about 20\%. While smaller grid number produces too rough features, resulting in poorer effect. For the hyperparameter $\epsilon$, its value is proportional to the number of positive grids generated by each instance. Too large $\epsilon$ is prone to overlap and occlusion of center regions between instances. Too small $\epsilon$ will result in too small response area on the feature map, and the generalization ability of the model is limited. Finally, we choose 1$\times$ grid numbers and $\epsilon=0.2$ to obtain the best trade-off between accuracy and speed. %which is mainly reflected that the offset is no longer easy to point to the valid location whose confidence exceeds the threshold. 

\vspace{-3mm}

\section{Conclusion}
This paper proposes to understand humans with point sets and center-based offsets, which results in a new framework, namely SMP, to solve the instance-aware multi-human parsing task in a single stage. Specifically, the feature of points in barycenters of human parts is utilized to generate the masks of the part instances. The offsets from the human center to part barycenters are used to unify the human instance. In order to enhance the representation of instance features for classification, we propose the Refined Feature Retain (RFR) module, which can utilize mask features to generate mask attention to guide feature extraction. For the problem of fashion classification errors with high inter-class similarity, we propose the Mask of Interest Reclassify (MIR) module, which employs the generated mask as the region of interest to refine the classification result. SMP has the advantages of fast inference, high accuracy and concise pipeline, which contribute to the human-centered research fields. 

%This paper proposes a new framework to solve the instance-aware multi-human parsing task, which can simultaneously complete the localization of human instances and its part instances, the mapping from human instances to part instances, and the fine generation of part instance masks in a single stage. The difficulty of mapping between human body and part instances can be greatly reduced by applying points to represent instances with different fine granularity. The point feature in the barycenter of the part instance is utilized to generate the mask of the part instance, which not only reduces the amount of calculation, but also improves the expression ability of the instance feature. In order to solve the problem of insufficient representation of instance features for classification, we propose Refined Feature Retain (RFR) module, which can utilize mask features generate mask attention to guide model extract features. For the problem of clothing classification errors with strong inter-class similarity, we propose the Mask of Interest Reclassify (MIR) module, which employs the generated mask as the region of interest to refine classification result. These designs together constitute an end-to-end trainable, single-stage instance-aware human semantic parser. It has the advantages of fast inference, high accuracy and small amount of calculation. We believe that it will make contributions to the future human-centered research fields. 

\bibliographystyle{named}
\bibliography{ijcai23}

\end{document}

\maketitle

This supplementary material provides the detailed information that the paper cannot show due to the length limit, including: 1) The performance comparison of our single stage method on the Densepose COCO dataset~\cite{RizaAlpGuler2018DensePoseDH}. 2) Visualization of multi-human parsing results compared with groundtruth on MHPv2.0~\cite{JianZhao2018UnderstandingHI}. 3) Some detailed information during training and inference.

\begin{figure}[h]
\centering
\includegraphics[width=1.0\columnwidth]{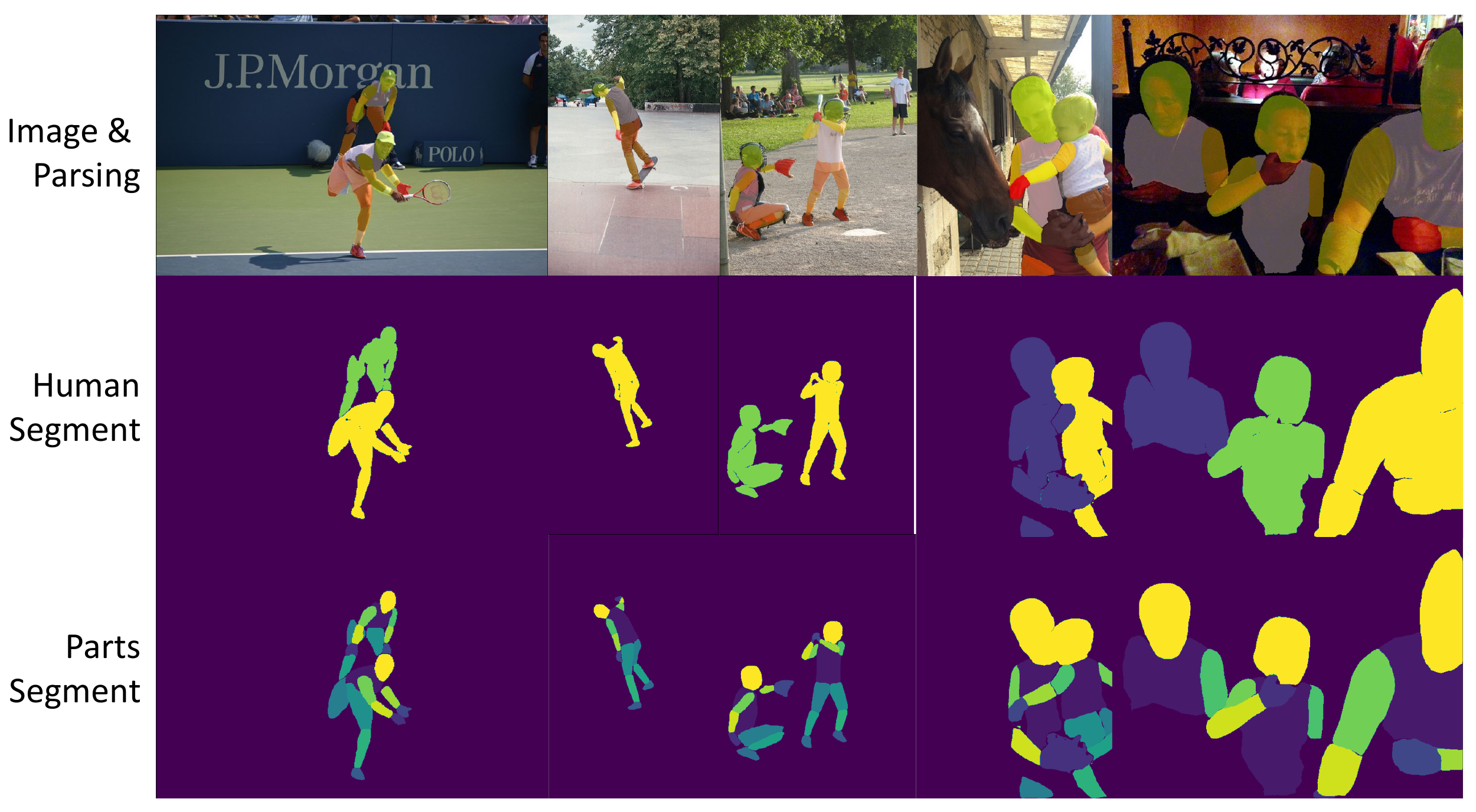} % Reduce the figure size so that it is slightly narrower than the column. Don't use precise values for figure width.This setup will avoid overfull boxes.
\caption{The visualization of the results on DensePose COCO.}
\label{The visualization of the resualts on DensePose COCO}
\end{figure}

\begin{figure}[h]
\centering
\includegraphics[width=1.0\columnwidth]{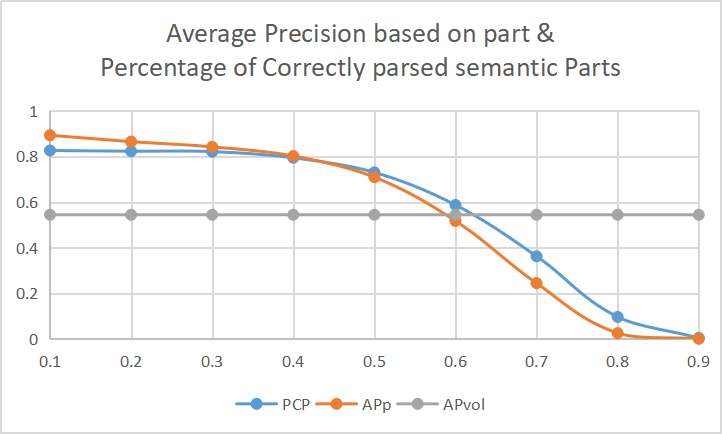} % Reduce the figure size so that it is slightly narrower than the column. Don't use precise values for figure width.This setup will avoid overfull boxes.
\caption{The values of AP$^{p}$ and PCP at different thresholds.}
\label{AP}
\end{figure}

\begin{figure*}[]
\centering
\includegraphics[width=2.0\columnwidth]{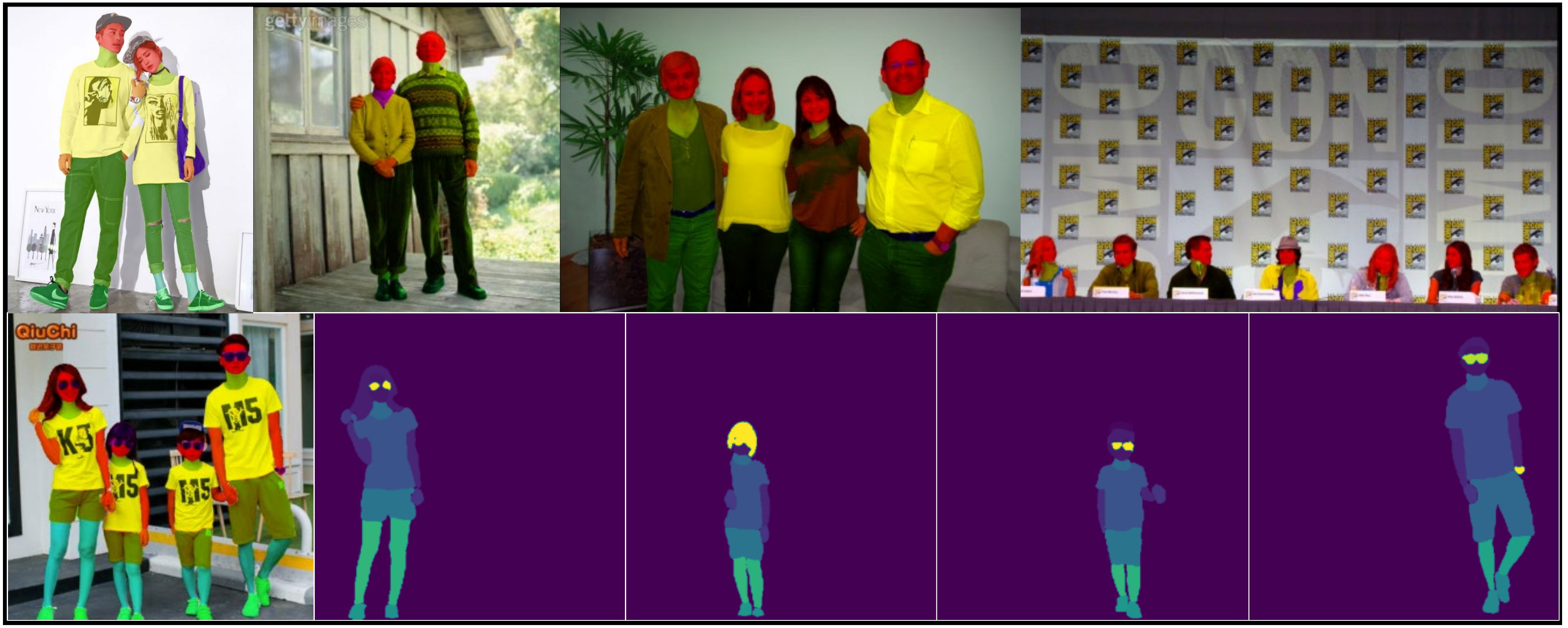} % Reduce the figure size so that it is slightly narrower than the column. Don't use precise values for figure width.This setup will avoid overfull boxes.
\caption{The visual results on MHPv2.0 including separated instances.}
\label{The visual instnace results on MHPv2.0 including separated instances.}
\end{figure*}

\begin{figure*}[]
\centering
\includegraphics[width=2\columnwidth]{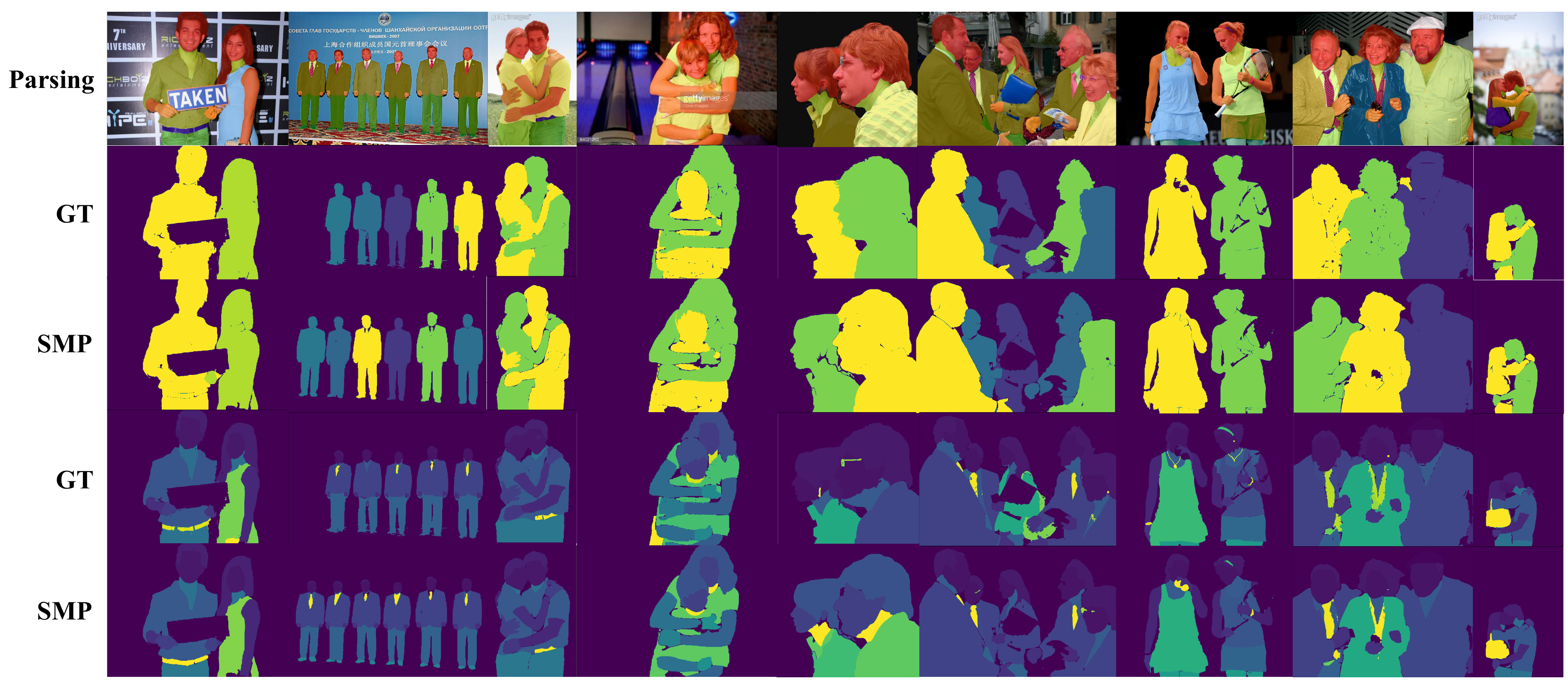} % Reduce the figure size so that it is slightly narrower than the column. Don't use precise values for figure width.This setup will avoid overfull boxes.
\caption{The visual results on MHPv2.0 compared with groundtruth.}
\label{The visual instnace results on MHPv2.0}
\end{figure*}

\section{The Performance on Densepose-COCO dataset}

We apply our method on the DensePose-COCO dataset. AP$^{p}_{50}$, AP$^{p}_{vol}$ and PCP$_{50}$ are also selected as metrics and the results are compared with other state of the art (SOTA) top-down and bottom-up methods. The backbone selected in the experiment is ResNet101 with deformable convolution~\cite{XizhouZhu2018DeformableCV}. Since the instance scales of the COCO dataset are small~\cite{LuYang2018ParsingRF}, we change the size of the grid in model to twice the original one.

\noindent\textbf{Dataset.} DensePose-COCO~\cite{RizaAlpGuler2018DensePoseDH} has 27,659 images (26,151/1,508 for train/test splits) gathered from COCO~\cite{TsungYiLin2014MicrosoftCC}. It annotates 14 human parts and 17 keypoints. 

\noindent\textbf{Metrics.}
As shown in paper, AP$^{p}$~\cite{JianZhao2018UnderstandingHI} calculates mean part instance level pixel intersection of union (IoU) of different semantic part categories within a human instance to determine if the human instance is a true positive. We choose the AP$^{p}_{50}$ and AP$^{p}_{vol}$ as the evaluation metrics. The former defines the instance whose IoU is larger than the threshold of 0.5 as the positive, and the latter is the average of AP$^{p}$ in the IoU threshold range from 0.1 to 0.9 in increments of 0.1. In addition, we also report the official metric, Percentage of Correctly parsed semantic Parts (PCP)~\cite{JianZhao2018UnderstandingHI}.

\begin{table}[t]
\caption{The results performed on Densepose-COCO dataset.}
\setlength{\tabcolsep}{4pt}
\begin{tabular}{lccc}
\hline
\textbf{Methods}      & \textbf{AP$^{p}_{50}$} & \textbf{AP$^{p}_{vol}$} & \textbf{PCP$_{50}$} \\ \hline
\textit{Top-Down}     &                 &                  &              \\ \hline
P-RCNN~(Yang et al. CVPR'18.)          & 43.5            & 53.1             & 51.8         \\
M-CE2P~(Ruan et al. AAAI'19.)                & 43.7            & 52.9             & 51.2         \\
RP-RCNN~(Yang et al. ECCV'20.)               & 48.5            & 54.4             & 51.1         \\ \hline
\textit{Bottom-Up}    &                 &                  &              \\ \hline
PGN~(Gong et al. ECCV'18.)       & 23.4             & 35.9      & 32.5        \\
NAN~(Zhao et al. IJCV'20.)  & 37.6            & 48.3             & 43.9         \\
DSPF~(Zhou et al. CVPR'21.) & 49.7            & \textbf{54.7}             & 52.8         \\ \hline
\textit{Single-stage} &                 &                  &              \\ \hline
SMP~(Ours)             &   \textbf{70.9}          &  54.4          &  \textbf{73.0}        \\ \hline
\end{tabular}
\captionsetup{font={small}}
\label{The results performed on Densepose COCO dataset.}
\end{table}

\noindent\textbf{Results.}
As shown in Tab.~\ref{The results performed on Densepose COCO dataset.}, our method achieves comparable performance with the SOTA methods on the DensePose-COCO. In AP$^{p}_{50}$, our model achieves 70.9\%, which is 21.2\% higher than the previous optimal model DSPF~\cite{TianfeiZhou2021DifferentiableMH}. In PCP$_{50}$, SMP reaches 73.0\%, which is also 20.2\% higher. For AP$^{p}_{vol}$, our method reaches 54.4\%, which is only 0.3\% lower than DSPF~\cite{TianfeiZhou2021DifferentiableMH}, but higher than other SOTA methods~\cite{LuYang2018ParsingRF}. From the results, we can see that our method has excellent performance on the Average Precision with 0.5 threshold, and the average performance is basically the same as the current optimal method. This indicates that our method has a higher lower limit than other methods, but it performs undesirably in the upper limit.

As shown in Fig.~\ref{AP}, when the threshold is smaller than 0.5, the AP$^{p}$ drops very slowly, while the threshold is larger than 0.5, the AP$^{p}$ drops rapidly. it means that our method can get a good result for most inputs. However, it is difficult to generate multi-human parsing whose IoU are above 0.8.

\section{Visualization of Multi-human Parsing Results}
In this appendix, we will show the visualization results of the model in detail, including the display by instance and the comparison with the ground truth. As shown in Fig.~\ref{The visual instnace results on MHPv2.0 including separated instances.}, We show the prediction results of different instances in the same image. In the first line, we display the results of combined multi-human parsing. In the second line, we provide the human parsing separate by instances. SMP does need neither bounding boxes~\cite{KaimingHe2017MaskR} to locate nor keypoints~\cite{TianfeiZhou2021DifferentiableMH} or edge~\cite{TaoRuan2019DevilIT} to match. The output results of SMP are directly human parsing of individual instances, as shown in the second line of Fig.~\ref{The visual instnace results on MHPv2.0 including separated instances.}.

In Fig.~\ref{The visual instnace results on MHPv2.0}, we provide the visualization of groundtruth and prediction of human instance level segmentation and those of part instance segmentation. Our method has achieved good results in the segmentation of two different granularity instances. Even in the face of complex occlusion or long-distance body parts, SMP can still properly match the human body and part instances.

\section{The Detailed Setting in Training and Inference}
Some hyperparametric settings involved in training and reasoning will be introduced in this appendix.

\noindent\textbf{Training.}
During training, our initial learning rate is set as 0.02, and the batch size per card is 4. The total batch size on 8 NVIDIA Tesla V100 is 32.%, and the average learning rate on each image is 0.0625. 

\noindent\textbf{Inference.}
During inference, we utilize $\theta_{c}$ as the threshold for localizing the barycenters of the human body in center heatmap $H_{C}$ with a value of 0.1. The threshold for localizing part instance $\theta_{p}$ has a value of 0.3. To avoid similar instances responding at the same location, we applied matrix NMS~\cite{XinlongWang2020SOLOv2DA} to filter the prediction of instances with large overlap with a threshold $\theta_{u}$ whose value is 0.1. 

And in the last of the appendix, we also provide the config file of our method in mmdetection style. It is very easy to understand, and is divided to model settings, dataset settings, optimizer settings and learning policy. 

\vspace{3cm}

\begin{lstlisting}[
    style       =   Python,
    caption     =   {\bf config.py},
    label       =   {config.py}
]
# model settings
model = dict(
    type='SingleStageInsParsingDetector',
    backbone=dict(
        type='ResNet',
        depth=101,
        num_stages=4,
        out_indices=(0, 1, 2, 3),
        frozen_stages=1,
        init_cfg=dict(type='Pretrained', checkpoint='torchvision://resnet101'),
        style='pytorch'),
    neck=dict(
        type='FPN',
        in_channels=[256, 512, 1024, 2048],
        out_channels=256,
        start_level=0,
        num_outs=5),
    mask_head=dict(
        type='SMP_head',
        num_classes=59,
        in_channels=256,
        stacked_convs=4,
        seg_feat_channels=512,
        strides=[8, 8, 16, 32, 32],
        scale_ranges=((1, 96), (48, 192), (96, 384), (192, 768), (384, 2048)),
        sigma=0.2,
        num_grids=[40, 36, 24, 16, 12],
        ins_out_channels=256,
        enable_center=True,
        enable_offset=True,
        enable_moi = True,
        mask_feature_head=dict(
            in_channels=256,
            feat_channels=128,
            start_level=0,
            end_level=3,
            out_channels=256,
            mask_stride=4,
            norm_cfg=dict(type='GN', num_groups=32, requires_grad=True)),
        loss_ins=dict(
            type='DiceLoss',
            use_sigmoid=True,
            loss_weight=3.0),
        loss_cate=dict(
            type='FocalLoss',
            use_sigmoid=True,
            gamma=2.0,
            alpha=0.25,
            loss_weight=1.0),
        loss_center=dict(
            type='FocalLoss',
            use_sigmoid=True,
            gamma=2.0,
            alpha=0.25,
            loss_weight=1.0),
        loss_offset=dict(
            type='SmoothL1Loss',
            beta=1. / 9, 
            reduction='mean', 
            loss_weight=10.0)),
    train_cfg = dict(),
    test_cfg = dict(
        nms_pre=500,
        ctr_score=0.1,
        cate_score_thr=0.3,
        mask_thr=0.5,
        cate_update_thr=0.1,
        update_thr=0.1,
        kernel='gaussian',  # gaussian/linear
        sigma=2.0,
        max_per_img=100,
        debug_heatmap=False)
    )
    

# dataset settings
dataset_type = 'MHP'
data_root = 'LV-MHP-v2/'
img_norm_cfg = dict(
    mean=[123.675, 116.28, 103.53], 
    std=[58.395, 57.12, 57.375], 
    to_rgb=True)
train_pipeline = [
    dict(type='LoadImageFromFile_mhp'),
    dict(type='LoadAnnotations_mhp', 
        with_mask=True, with_seg=True ,
        with_mhp_parsing=True, 
        with_keypoints=True),
    dict(type='Resize_Parsing',
         img_scale=[(1333, 800), (1333, 768), (1333, 736),
                    (1333, 704), (1333, 672), (1333, 640)],
         multiscale_mode='value',
         keep_ratio=True),
    dict(type='RandomFlip_Parsing', 
        flip_ratio=0.5, dataset='mhp'),
    dict(type='Normalize', **img_norm_cfg),
    dict(type='Pad_Parsing', size_divisor=32),
    dict(type='DefaultFormatBundle_Parsing'),
    dict(type='Collect', keys=['img', 
                            'gt_bboxes', 
                            'gt_labels', 
                            'gt_masks', 
                            'gt_parsing', 
                            'gt_semantic_seg', 
                            'gt_keypoints'],
                         meta_keys=('filename', 
                            'ori_shape', 
                            'img_shape', 
                            'pad_shape',
                            'scale_factor', 
                            'flip', 
                            'img_norm_cfg')),
]
test_pipeline = [
    dict(type='LoadImageFromFile_mhp'),
    dict(
        type='MultiScaleFlipAug',
        img_scale=(1333, 800),
        flip=False,
        transforms=[
            dict(type='Resize_Parsing', keep_ratio=True),
            dict(type='RandomFlip_Parsing', dataset='mhp'),
            dict(type='Normalize', **img_norm_cfg),
            dict(type='Pad_Parsing', size_divisor=32),
            dict(type='ImageToTensor', keys=['img']),
            dict(type='Collect', keys=['img']),
        ])
]
data = dict(
    samples_per_gpu=4,
    workers_per_gpu=4)
# optimizer
optimizer = dict(type='SGD', 
                  lr=0.02, 
                  momentum=0.9, 
                  weight_decay=0.0001)
optimizer_config = dict(
                    grad_clip=dict(
                        max_norm=35, 
                        norm_type=2))
# learning policy
lr_config = dict(
    policy='step',
    warmup='linear',
    warmup_iters=500,
    warmup_ratio=0.02,
    step=[8, 11])
runner = dict(type='EpochBasedRunner', max_epochs=12)
\end{lstlisting}

\bibliographystyle{named}
\bibliography{ijcai23}

%\bibliography{ijcai23}